\documentclass[letterpaper, 10 pt, conference]{ieeeconf}
\usepackage[english]{babel}

\usepackage{cite}
\usepackage{multicol}
\usepackage{amsmath}
\usepackage{amsfonts}
\usepackage{graphicx}
\usepackage{amssymb}
\usepackage{algorithm}
\usepackage{algpseudocode}
\usepackage[bookmarks=true,hidelinks]{hyperref}
\usepackage{subcaption}
\usepackage{booktabs}
\usepackage{todonotes}
\usepackage{multirow}
\usepackage{lipsum}  
\usepackage{bbm}

\newcommand{\successnode}{\textsc{Success}}

\newcommand{\rollout}{\tau}

\DeclareMathOperator*{\avg}{avg}

\let\oldtwocolumn\twocolumn
\renewcommand\twocolumn[1][]{%
    \oldtwocolumn[{#1}{
    \begin{center}
    \vspace{-20pt}
           \includegraphics[width=\textwidth]{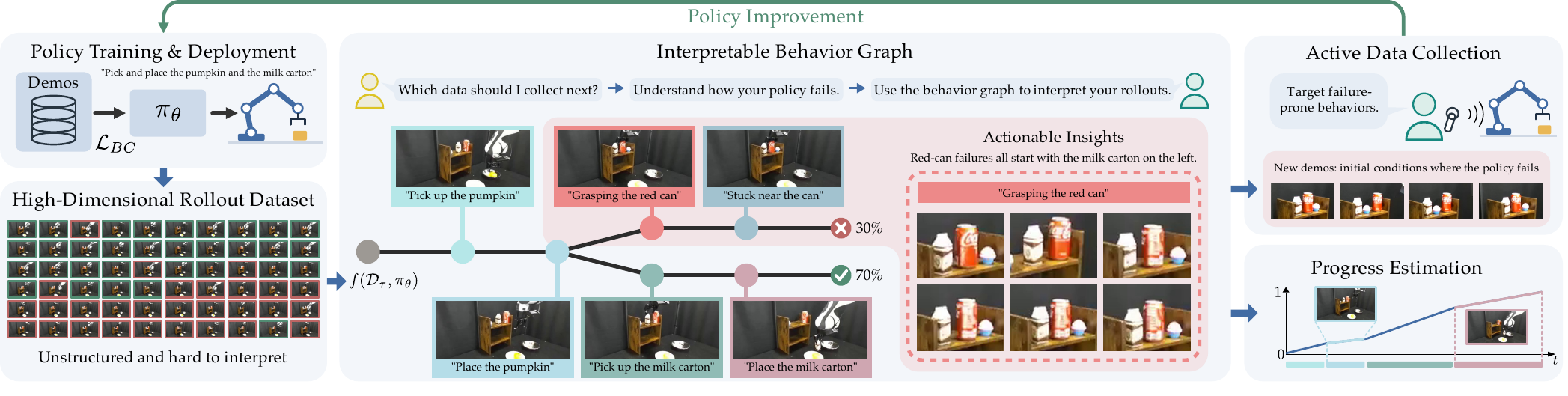}
           \captionof{figure}{\small How do robot policies fail during deployments, and which data should we collect to improve them? Raw policy rollouts that deployments produce are hard to interpret and offer limited insights. We introduce behavior graphs, which compress policy rollouts into an interpretable graph. Data collectors and researchers can use this structure to identify failure modes and guide targeted data collection, closing the policy improvement loop. Behavior graphs also enable state-of-the-art progress estimation without any additional model training.}
           \label{fig:title_figure}
        \end{center}
    }]
}

\usepackage{cleveref}
\Crefname{figure}{Fig.}{Figs.}
\Crefname{table}{Tab.}{Tabs.}
\Crefname{section}{Sec.}{Secs.}
\Crefname{subsection}{Sec.}{Secs.}
\Crefname{subsubsection}{Sec.}{Secs.}
\Crefname{equation}{Eq.}{Eqs.}

\begin{document}
\nocite{IEEEexample:BSTcontrol}
\title{\LARGE \bf Behavior Graphs: Unsupervised Graph Structures for Interpreting Robot Deployment Data}



\newcommand{\authorhref}[3][black]{\href{#2}{\textcolor{#1}{#3}}}
\author{
    \bfseries
    \authorhref{http://erikbr01.github.io/}{Erik Bauer}\textsuperscript{1},\,\,
    \authorhref{https://jakob-thumm.com/}{Jakob Thumm}\textsuperscript{1},\,\,
    \authorhref{http://agiachris.github.io/}{Christopher Agia}\textsuperscript{1},\,\,
    \authorhref{https://rohansinha.nl/}{Rohan Sinha}\textsuperscript{1}\\ \bfseries
    \authorhref{https://hamrel-cxu.github.io/}{Chen Xu}\textsuperscript{3},\,\,
    \authorhref{https://harukins.github.io/}{Haruki Nishimura}\textsuperscript{3},\,\,
    \authorhref{https://web.stanford.edu/~bohg/}{Jeannette Bohg}\textsuperscript{1},\,\,
    \authorhref{https://profiles.stanford.edu/marco-pavone}{Marco Pavone}\textsuperscript{1,2},\,\,
    \authorhref{https://mashaitkina.weebly.com/}{Masha Itkina}\textsuperscript{3}\\\vspace{-8pt}\\
    \textsuperscript{1}\href{https://www.stanford.edu/}{Stanford University},
    \textsuperscript{2}\href{https://www.nvidia.com/en-us/research/}{NVIDIA Research},
    \textsuperscript{3}\href{https://www.tri.global/}{Toyota Research Institute}\\\vspace{-8pt}\\
    \small\url{https://behavior-graphs.com}
}%

\maketitle







\begin{abstract}
Data flywheels promise continuous improvement via insights from deployment data. Understanding how the deployed model succeeds and fails can steer subsequent data collection towards patching failures and reinforcing successes. However, robot deployments create large volumes of high-dimensional rollout data, which is challenging to understand and confidently ground data collection decisions in. To make deployment data easier to analyze and act on, we propose behavior graphs, which compress policy rollouts into a single interpretable, directed graph. We construct the graphs by learning to identify similar behavior modes across rollouts, connecting the behaviors temporally, and finally linking them to outcomes. In a user study on three real-world tasks, where participants analyzed rollouts to choose a data collection strategy, 16 of 18 preferred behavior graphs to rollout videos, reaching comparable conclusions with 40.3\% lower perceived temporal demand and 37.6\% lower effort. Beyond interpretability for deployment data, behavior graphs are capable progress predictors: across 43 datasets, they improve the rank correlation between predicted progress and true temporal order by 22.9\% and reduce mean absolute error by 53.5\% relative to the strongest baseline. As the number of deployed robots grows, behavior graphs lend structure to high-dimensional policy rollouts and make deployment data easier to interpret and reason about.
\end{abstract}

\section{Introduction}

Imitation learning has become a standard approach for training robotic manipulation policies~\cite{chi_diffusion_2023,zhao_learning_2023}. As policies move toward real-world deployment, improving them after initial training relies on the \emph{data flywheel}: deploy the policy, observe its rollout behavior, collect new demonstrations that address observed shortcomings, and retrain. Closing this loop effectively is difficult. Operators must identify not just that a policy fails, but which execution strategies are responsible, whether those failures stem from missing or low-quality training data, and what new demonstrations would meaningfully change the policy's behavior distribution. Without a structured account of policy behavior, these decisions reduce to intuition. This raises two problems as robots move from labs to deployments: rollout data quickly outgrows what operators can review, and intuition is hard to communicate, compare, or verify. Closing the data flywheel requires methods that condense deployment data into a shared basis for understanding and reasoning about policy behavior.

Most existing works that attempt to learn from policy rollouts focus on direct policy improvement rather than generating interpretable insights. Data curation methods identify which existing demonstrations to discard~\cite{agia_cupid_2025, zhang_scizor_2025, chen_sarm_2025, intelligence2025pi06vlalearnsexperience}, but not which demonstrations to collect next. While red-teaming approaches~\cite{karnik2024embodied, majumdar2025predictive} can generate valuable insights about failure modes of a policy, they rely on actively steering policy inputs to cause failures, which is useful for finding weaknesses in a controlled setting, but may overly represent pessimistic scenarios compared to typical deployment settings. Active data collection approaches such as DAgger~\cite{kelly2019hgdaggerinteractiveimitationlearning, ross_reduction_2011} are effective, but expensive to scale, as they require constant monitoring and human experts to collect recovery data. None of these methods helps operators interpret and act on autonomous rollout data from robot deployments.

We address this gap with \emph{behavior graphs}, a representation that compresses policy rollouts into an interpretable directed graph~(\Cref{fig:title_figure}). Nodes in a behavior graph correspond to semantic behavior modes, which are connected by edges that represent the transition probabilities between different behaviors and rollout outcomes. Construction is unsupervised: we cluster policy embeddings of rollout segments from all rollouts into shared behavior modes, which form the graph’s nodes. Directed edges connect each behavior to its successors and to rollout outcomes, weighted by the empirical transition probabilities observed across rollouts. This structure lets operators trace where, when, and how a policy fails, and steer data collection accordingly. In a user study where participants chose data collection strategies using either behavior graphs or rollout videos, behavior graphs lower perceived temporal demand by 40.3\% and effort by 37.6\%, without changing the conclusions participants reach. Furthermore, behavior graphs serve as a strong representation for progress estimation: across 43 datasets spanning simulation and the real world, they outperform specialized progress estimators, both learned and built on vision-language models (VLMs).

Our contributions are threefold:
\begin{itemize}
    \item Behavior graphs, an unsupervised representation of policy rollouts that clusters policy embeddings of rollout segments into behaviors and links them to each other and to rollout outcomes through empirical transition probabilities.
    \item A user study with three real-world tasks in which 16 of 18 participants prefer the graph over videos of rollouts, and perceived temporal demand and effort drop by 40.3\% and 37.6\% while data collection decisions remain comparable.
    \item A method for progress estimation using behavior graphs that achieves state-of-the-art results across 43 datasets, improving value-order correlation by 22.9\% and reducing mean absolute error by 53.5\% relative to the strongest baseline.
\end{itemize}

\section{Related Work}

\subsubsection{Identifying behavior from policies and trajectories} Hidden activations of vision-language-action (VLA) models encode executed behaviors and can be used for steering~\cite{swann_sparse_2026, jin_eventgrounded_2026}, and unsupervised subtask segmentation~\cite{blank2026sparc, yang_unsupervised_2025, wang_grounding_2024} has been applied to skill transfer and retrieval~\cite{sharma_robosubtasknet_2026, memmel_strap_2025, chen_backbone_2025}. These works recover behavioral structure from policies and trajectories, but do not connect the identified behaviors to each other, nor use them to inform what data to collect next.

\subsubsection{Representing policies as graphs} Zahavy et al.~\cite{zahavy_graying_2016} and Topin et al.~\cite{topin_generation_2019} both abstract low-dimensional reinforcement learning policies into behavioral Markov decision processes, clustering deep Q-network representations in one case and grouping states by feature importance in the other. Graph structures can also be utilized as controllers. Behavior trees~\cite{iovino_survey_2022} are authored by hand rather than discovered from existing trajectories, and define a control flow the robot is expected to follow. Pan et al.~\cite{pan2026emergentneuralautomatonpolicies} extract automaton structure from visuomotor trajectories to construct control flows that define a policy. These methods prescribe policies rather than describe the actual observed rollout behavior, and neither close the loop to active data collection nor perform progress estimation.

\subsubsection{Policy improvement from deployment data} Curation methods use policy rollouts to score and reweight existing training data, whether through influence functions, self-supervised criteria, or learned value functions~\cite{agia_cupid_2025, zhang_scizor_2025, intelligence2025pi06vlalearnsexperience}, but do not reason about which data to collect next. Red-teaming approaches~\cite{karnik2024embodied, majumdar2025predictive} actively attempt to break the policy by assuming control over its observations and actively steering them towards failure-inducing regions, which is not representative of the typical deployment setting for robots that we consider. Variants of DAgger~\cite{ross_reduction_2011, kelly2019hgdaggerinteractiveimitationlearning} rely on experts monitoring policies for interventions and collecting failure recovery data, which is effective, but costly to scale. Zha et al.~\cite{zha_guiding_2025} guide collection by measuring performance scaling over coarse, manually defined environmental factors.

\subsubsection{Progress estimation} Progress estimators support rollout analysis and reward modeling. Stage-conditioned estimators are interpretable, but require hand-annotated subtasks~\cite{chen_sarm_2025, kim2025subtask}. Generalist progress estimators such as Robometer~\cite{liang_robometer_2026} and TOPReward~\cite{chen_topreward_2026} need no task-specific training, but their accuracy varies widely across domains. Existing estimators thus trade annotation effort for precision.

\section{Methodology}

We first introduce methodology for unsupervised behavior graph construction, followed by how graphs can be used for active data collection and progress estimation.

\subsection{Behavior Graph Construction}

\begin{figure*}
  \vspace{10pt}
  \begin{center}
    \includegraphics[width=\linewidth]{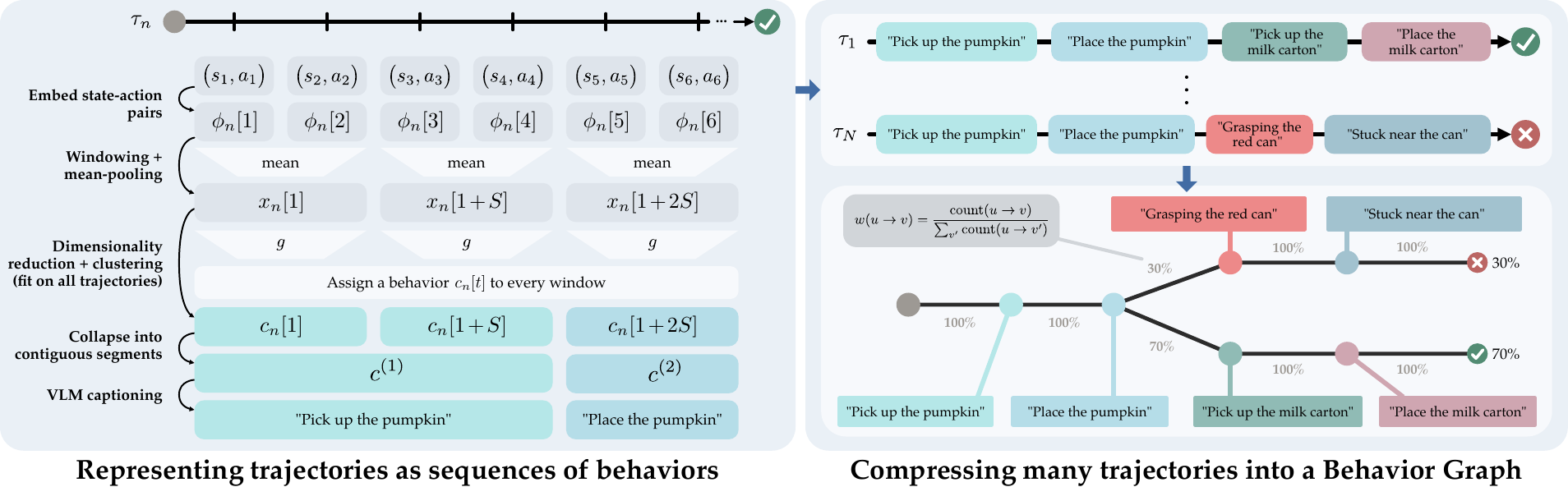}
  \end{center}
  \caption{\small Behavior graph construction. Left: Each trajectory’s state-action pairs are embedded, mean-pooled over temporal windows, and assigned to behavior modes by dimensionality reduction and clustering fit on all trajectories. Consecutive windows with the same mode are collapsed into segments, and a VLM captions each mode. Right: The resulting behavior sequences from all trajectories are merged into a single directed graph from a start node to success and failure outcomes, with edges weighted by empirical transition probabilities.}
  \label{fig:behavior_graph_construction}
\end{figure*}

We are given a set of trajectories $\mathcal{D}_\tau = \{\tau_1, \ldots, \tau_N\}$, where each trajectory $\tau_n = \big((s_1, a_1), \ldots, (s_{T_n}, a_{T_n})\big)$ is a sequence of state-action pairs $(s_t, a_t) \in \mathcal{S} \times \mathcal{A}$ of length $T_n$, and we write $\tau_n[t]$ for its $t$-th element. Trajectories are either policy rollouts, whose actions are sampled from a policy $\pi_\theta(a \mid s)$ with parameters $\theta$, or training demonstrations, whose actions are labeled by an expert. Each trajectory is annotated with a binary outcome $o(\tau) \in \{\textsc{Success}, \textsc{Failure}\}$. Finally, we assume access to an embedding function $f\colon \mathcal{S} \times \mathcal{A} \to \mathbb{R}^d$. 

Our aim is to discover a finite set of behaviors $\mathcal{C} = \{c_1, \ldots, c_K\}$ from $\mathcal{D}_\tau$, represent each trajectory $\tau_n$ as a sequence $\sigma_n \in \mathcal{C}^{L_n}$ of $L_n$ behaviors, and compress all sequences into an interpretable directed graph $G = (V, E, w)$ whose nodes $V$ are the behaviors and outcomes, with edges $E$ weighted by $w$ (\Cref{fig:behavior_graph_construction}).

\begin{figure}
\vspace{10pt}
    \centering
    \includegraphics[width=0.95\linewidth]{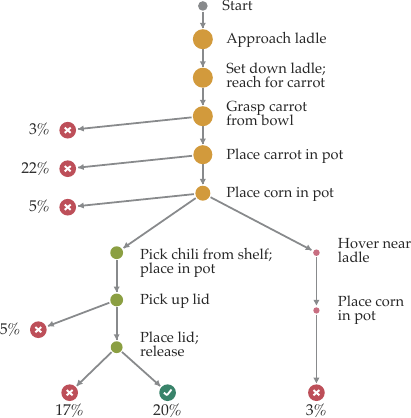}
    \caption{\small Behavior graph for the cooking task of our user study~(\Cref{sec:user_study}). The larger the node, the higher the visitation count. Node colors indicate success likelihood (increasing from red to green), while all terminal nodes are annotated with their likelihoods. Branches with less than 2\% likelihood are pruned to simplify visualization.}
    \label{fig:set_table_behavior_graph}
\end{figure}

\paragraph{Embedding trajectories}
We embed every timestep of every trajectory as $\phi_n[t] = f(\tau_n[t])$. The choice of $f$ determines the notion of similarity the graph is built on. Unless stated otherwise, we use policy embeddings, where $f$ returns the output of the observation encoder of $\pi_\theta(a \mid s)$.

\paragraph{Temporal aggregation}
Since behaviors span multiple timesteps and per-timestep embeddings are noisy, we smooth each embedding sequence $\phi_n$ with a sliding-window mean filter of width $W$ and stride $S$, and write $x_n[t] \in \mathbb{R}^d$ for the window starting at timestep $t$.

\paragraph{Dimensionality reduction and clustering}
To facilitate clustering, we map each window to a lower-dimensional space with a reduction $g\colon \mathbb{R}^d \to \mathbb{R}^{d_\text{low}}$, fit on the windows of all trajectories. Unless otherwise stated, we use UMAP as dimensionality reduction. We then run $K$-means on the reduced windows $g(x_n[t])$, yielding $K$ clusters with centroids $\mu_1, \ldots, \mu_K \in \mathbb{R}^{d_\text{low}}$. Each cluster defines a behavior $c_k$, giving the set of behaviors $\mathcal{C} = \{c_1, \ldots, c_K\}$. Each window $x_n[t]$ is assigned the behavior $c_n[t]$ based on the cluster assignment in the reduced space,
\begin{equation}
  c_n[t] = c_{k^*}, \quad k^* = \arg\min_k \lVert g(x_n[t]) - \mu_k \rVert_2.
  \label{eq:assignment}
\end{equation}

\paragraph{Segments}
Since behaviors typically span multiple windows, consecutive windows are often assigned the same behavior. We define a \emph{segment} as a contiguous sequence of windows that have the same behavior assigned to them. Ordering the $L_n$ segments of $\tau_n$ in time, the trajectory is reduced to its sequence of behaviors
\begin{equation}
  \sigma_n = \big(c^{(1)}, \ldots, c^{(L_n)}\big), \qquad c^{(l)} \neq c^{(l+1)},
  \label{eq:sequence}
\end{equation}
where $c^{(l)}$ is the behavior of the $l$-th segment.

\paragraph{Graph construction}
The behavior sequences define a directed graph $G = (V, E,w)$ over behaviors and outcomes, with node set $V = \{\textsc{Start}\} \cup \mathcal{C} \cup \{\textsc{Success}, \textsc{Failure}\}$. Each sequence $\sigma_n$ contributes an edge $\textsc{Start} \to c^{(1)}$, the inter-behavior edges $c^{(l)} \to c^{(l+1)}$, and a terminal edge $c^{(L_n)} \to o(\tau_n)$ into the trajectory's outcome. Since no two consecutive elements of $\sigma_n$ are equal, $E$ contains no self-loops. In addition, every edge carries the probability of taking that transition, estimated from the counts \(\mathrm{count}(u \to v)\) accumulated over all trajectories and normalized over the outgoing edges of its source:
\begin{equation}
  w(u \to v) = \frac{\mathrm{count}(u \to v)}{\sum_{v'} \mathrm{count}(u \to v')}.
  \label{eq:edge_weights}
\end{equation}

These weights assume that transitions are first-order Markovian, and whether that assumption holds is not always clear. We have found prefix trees to be a viable alternative: a node is a path prefix, so execution modes that visit the same behaviors in different orders stay distinct and no Markov assumption is needed. However, prefix trees give up the graph's bound on node count, growing instead with the number of distinct orderings the policy exhibits. Nothing in the remainder of our methodology relies on the Markov assumption, so the choice between the two reduces to a preference in visualization.

\paragraph{VLM annotation}
Captioning the discovered behaviors \(\mathcal{C}\) makes it easier to understand the graph at a glance, rather than having to inspect all behaviors to understand their meaning. To enable automatic captioning by VLMs, we utilize contact sheets. A contact sheet for a behavior \(c \in \mathcal{C}\) is a \(q \times m\) grid of images, which samples \(q\) segments that were assigned to \(c\), and represents each segment with \(m\) uniformly sampled images. The VLM receives the contact sheet together with a text description of the graph topology, and optionally, a task instruction as specified by the user. In a first pass, the VLM generates a caption \(\ell_c\) for every behavior. In a second pass, it reflects on the captions, evaluates global coherency, and corrects them if necessary.

\subsection{Use Cases}
\label{subsec:use_cases}
 
We consider two use cases for behavior graphs: active data collection and progress estimation. For data collection, we read quantities directly off the graph, while for progress estimation, we fit a lightweight model on top of it.
 
\paragraph{Active data collection}
To support failure analysis, behavior graphs expose distinct execution modes and their outcomes. Two quantities follow directly from the graph and let operators reason about failure at the level of individual behaviors or entire execution strategies. The \emph{success likelihood} of a behavior \(c\) is the fraction of trajectories that visited \(c\) and ended in success:
\begin{equation}
  p_\text{succ}(c) = \frac{\bigl|\{n : c \in \sigma_n,\; o(\rollout_n) = \successnode\}\bigr|}{\bigl|\{n : c \in \sigma_n\}\bigr|}.
  \label{eq:psucc}
\end{equation}
This isolates failure-prone behaviors independently of where in the execution they occur. When read together with the visitation count \(\bigl|\{n : c \in \sigma_n\}\bigr|\), it identifies clear data collection targets: behaviors that are both frequently visited and result in failure. Visually, we show the success likelihood via a color gradient for nodes in the graph, ranging from red for \(p_\text{succ} = 0\) to green for \(p_\text{succ} = 1\)~(\Cref{fig:set_table_behavior_graph}), and node radius encoding the visitation count. Furthermore, we can consider likelihoods of entire paths \(\rho = \bigl(\textsc{Start}, c^{(1)}, \ldots, c^{(L)}, o\bigr)\), computed as the product of their edge weights,
\begin{equation}
  P(\rho) = \prod_{l=0}^{L} w\bigl(c^{(l)} \to c^{(l+1)}\bigr),
  \label{eq:pathprob}
\end{equation}
with \(c^{(0)} = \textsc{Start}\) and \(c^{(L+1)} = o\). On a prefix tree, both quantities are read per branch instead. Writing \(\rho \preceq \sigma_n\) when the behaviors along branch \(\rho\) form a prefix of \(\sigma_n\), and \(N(\rho) = \bigl|\{n : \rho \preceq \sigma_n\}\bigr|\) for the number of rollouts that followed it, a branch's likelihood is \(P(\rho) = N(\rho)/N\) and its success likelihood is \(p_\text{succ}(\rho) = \bigl|\{n : \rho \preceq \sigma_n,\; o(\rollout_n) = \successnode\}\bigr| / N(\rho)\). By default, we prune branches whose likelihood falls below 2\% to make prefix trees easier to interpret at a glance.

Ranking paths by \(P(\rho)\) surfaces the policy's dominant execution strategies, so that operators can see which failure modes occur most often and allocate collection budgets accordingly. The added temporal context of analyzing paths instead of individual behaviors can be helpful in cases where the root cause of a failure lies in a behavior that is upstream in the graph rather than in the behavior that occurs immediately before failure.

Furthermore, users can interact with the graphs: they can click on nodes or transitions to see videos of all episodes that went through the node or transition, starting from the point in time where the behavior or the behavioral transition occurred. Thus, users can pinpoint high-level failures by considering either the node- or path-level failure candidates, and then analyzing the rollouts that have been grouped into that particular failure-prone behavior mode.
 
\paragraph{Progress Estimation}
Estimating how far a rollout progressed through its task can be useful for analyzing deployments retroactively. Such estimators are typically learned end-to-end from trajectories and progress labels. SARM~\cite{chen_sarm_2025} instead conditions progress on which subtask a rollout occupies: each subtask is assigned a share of total progress proportional to its average duration, and progress increases linearly within it. SARM uses this formulation to construct training labels from hand-annotated subtasks and then trains a neural reward model to predict them. We adopt the same formulation but differ in two ways: the subtasks are the behaviors the graph discovers, removing the annotation requirement, and the formulation itself serves as the estimator, without training a network.

Concretely, each behavior \(c\) is associated with a slope \(\alpha_c\) and an intercept \(\beta_c\), fit on successful rollouts. The slope is the average fraction of a trajectory's windows assigned to \(c\),
\begin{equation}
  \alpha_c = \avg_n \, \avg_t \, \mathbbm{1}\bigl[c_n[t] = c\bigr],
  \qquad \textstyle\sum_{c \in \mathcal{C}} \alpha_c = 1.
  \label{eq:stage_span}
\end{equation}
Ordering the behaviors \(c_k\) by their mean normalized time of occurrence, the intercepts accumulate the slopes of all preceding behaviors, \(\beta_{c_1} = 0\) and \(\beta_{c_k} = \beta_{c_{k-1}} + \alpha_{c_{k-1}}\). To score a new trajectory, we assign its windows to behaviors via \Cref{eq:assignment}, group them into segments, and predict progress \(\hat{p} = \beta_c + \alpha_c \bar{t}\) within each segment of behavior \(c\), where \(\bar{t} \in [0, 1]\) is the normalized time elapsed within the segment.

\section{Experiments}

\begin{figure*}[!t]
    \centering
    \includegraphics[width=\linewidth]{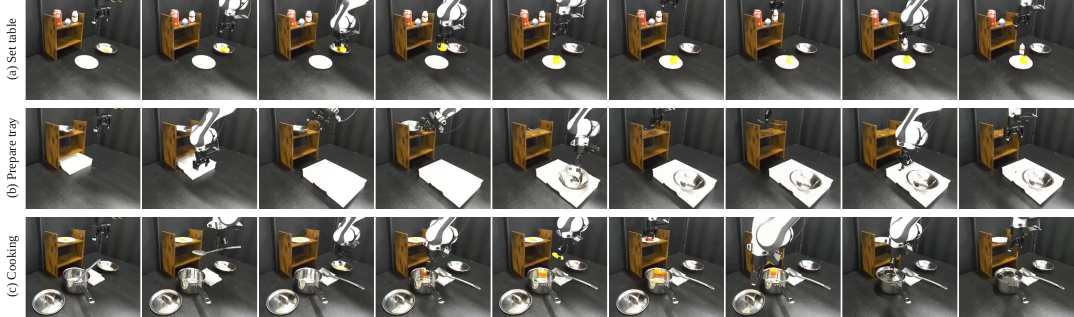}
    \caption{\small Tasks for the user study, in which participants analyze policy rollouts to choose which demonstrations to collect next. Keyframes of a successful rollout from left to right. (a) \emph{Set table}: places a pumpkin, then a milk carton from the shelf; the training data omits the carton's leftmost position. (b) \emph{Prepare tray}: slides out a tray, places a bowl from the top or bottom shelf onto it, and grasps a fork, requiring precise grasps. (c) \emph{Cooking}: a long-horizon task that clears a ladle, adds a corn, carrot, and chili to the pot, and closes the lid.
}
    \label{fig:user_study_tasks}
\end{figure*}

\begin{figure*}[!t]
    \centering
    \includegraphics[width=\linewidth]{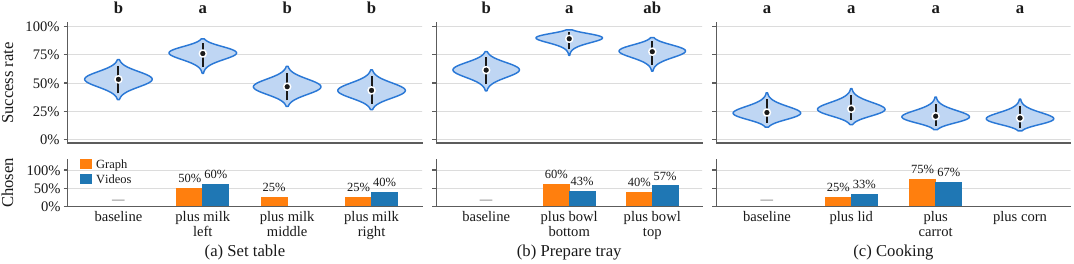}
    \caption{\small 
    Targeted data helps only when it addresses the policy's most frequent failure mode. Estimated success rates~\cite{barreiros_careful_2026} of the baseline policy and of policies retrained with 30 additional demonstrations for each candidate data-collection strategy, on three real-world DROID tasks. In {\em Set table\/}, only demonstrations with the milk carton on the left (the position missing from the baseline training data) improve performance. In {\em Prepare tray\/}, bottom-shelf bowl demonstrations improve over the baseline. In {\em Cooking\/}, no single-subtask strategy makes a significant difference. Violins show the Bayesian posterior over each policy's success rate (uniform Beta prior). Dots and bars mark the posterior mean and 95\% credible interval. Letters above each column~\cite{piepho2004algorithm} summarize pairwise significance tests within a task~\cite{snyder2025your}: two policies that share a letter are not significantly different; policies with no letter in common are.
    Tests follow~\cite{snyder2025your} with family-wise error rate controlled at 5\%, and 60 policy rollouts to estimate each posterior. Below, histograms show the strategies participants chose with the graph and with the video browser, which are similar.  
}
    \label{fig:overall}
\end{figure*}

\begin{figure*}
    \centering
    \includegraphics[width=\linewidth]{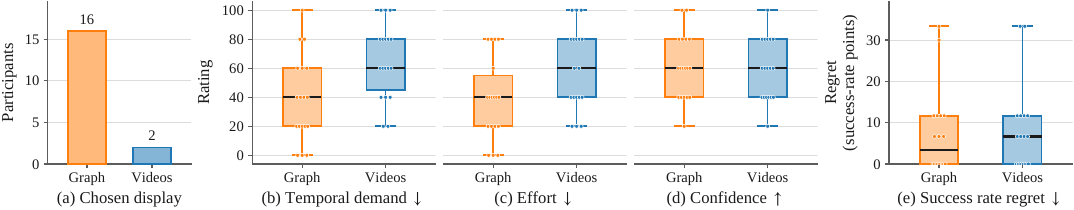}
    \caption{\small
    Behavior graphs make rollout analysis less demanding without changing the decisions users reach. Responses from 18 participants.  
    (a)~When free to choose a display for the third task, 16 of 18 participants picked the behavior graph over the video browser. (b-d)~Self-reported ratings (0-100, in steps of 20) after analyzing a task with each display: the graph lowers perceived temporal demand~(b) and effort~(c), while confidence in the chosen data-collection strategy is unchanged~(d). Boxes span the interquartile range with the median marked, whiskers span min/max values; dots are individual ratings. 
    (e) The success rate regret (i.e., the gap between the best retrained policy in \Cref{fig:overall} and the one the participant selected) 
    is similar for graph and videos. Arrows indicate whether higher~($\uparrow$) or lower~($\downarrow$) is better. \Cref{tab:study-adjusted} reports the corresponding effects adjusted for task, stage, and experience.
    }
    \label{fig:user_study_results}
\end{figure*}

We evaluate behavior graphs for active data collection and progress estimation, and ablate the graph construction methodology. We investigate three questions:
\textbf{Q1:} How do behavior graphs impact failure analysis and data collection decisions? (\Cref{sec:user_study})
\textbf{Q2:} How do behavior graphs compare to state-of-the-art progress estimators? (\Cref{sec:progress})
\textbf{Q3:} Which graph construction choices matter? (\Cref{sec:ablation})

\subsection{Behavior Graphs for Active Data Collection (Q1)}
\label{sec:user_study}
In a user study with 18 participants on three real-world tasks on a DROID~\cite{khazatsky2024droid} setup, participants analyze policy rollouts to select data collection strategies. We measure both how demanding the analysis is and how good the resulting decision is. Participants rate their perceived temporal demand, effort, and confidence. To score their chosen data collection strategy, we collected 30 additional demonstrations for each candidate strategy in advance, retrained the policy on each, and measured its success rate. Data collection choice are then scored by regret \(r = \mathrm{SR}_{\max} - \mathrm{SR}_{\text{chosen}}\), which is the gap of the achieved success rate to the success rate of the best possible option the participant could have chosen. 

Each participant analyzes 60 rollouts of each of the three tasks, in random task order. The first two are shown either as a behavior graph or in a video browser that displays an external and wrist camera view of six rollouts at a time, filterable by success or failure. For the third task, participants choose their preferred visualization, which we take as the preference measure.

\subsubsection{Real-World Tasks}
We choose three tasks that each expose a common source of policy failure (\Cref{fig:user_study_tasks}). \emph{Set table} targets missing initial conditions: the training data covers every shelf arrangement except those with the milk carton on the left. \emph{Prepare tray} targets precision: failures stem largely from the precise gripper placement the bowl and fork grasps require, and occur slightly more often for the bottom-shelf bowl, whose grasp is harder given the robot's kinematics; the non-prehensile tray slide additionally stresses subtask segmentation. \emph{Cooking} has a long horizon, which makes it more challenging to analyze via raw videos.

The candidate data collection strategies are the possible milk carton positions in \emph{set table}, the bowl's shelf height in \emph{prepare tray}, and individual subtasks in \emph{cooking}. Baseline policies are trained on 120 demonstrations for \emph{set table} and \emph{prepare tray}, and 200 for \emph{cooking}. \Cref{fig:overall} reports success rate from training on baseline data plus 30 added demonstrations for each data collection strategy.

\subsubsection{Analysis}

All analyses use the first two randomized stages of the study from 18 participants; the third stage contributes only the preference measure. For each outcome, we fit a linear mixed-effects model with fixed effects for visualization, task, and study stage, self-reported data-collection experience as a linear 0--4 score, and a random intercept per participant that captures differences in how participants use the rating scale.


We treat the six-point 0--100 ratings as continuous and report estimated marginal means (EMMs) per visualization, together with their difference, the graph effect, and its 95\% credible interval (CrI). Each EMM averages over task, study stage, and experience, with participant intercepts set to zero.



\subsubsection{Results}

Behavior graphs make rollout analysis faster and easier without changing what users conclude~(raw ratings in \Cref{fig:user_study_results}, EMMs in \Cref{tab:study-adjusted}). 
In particular, the graph reduces perceived temporal demand by 26.1 points and effort by 22.9 points compared to videos (40.3\% and 37.6\% relative decrease).  
User confidence in the selected data collection strategy and the resultant success rate regret (averaged across tasks) show no reliable difference, indicating that the graph incurs negligible information loss relative to video data for the purposes of active data collection.
Overall, 16 of 18 participants (89\%) chose the graph for the self-selected third task, indicating strong user preference. 


\begin{table}[t]
  \centering
  \caption{\small Effect of the behavior graph after adjusting for task, study stage, participant experience, and individual differences in scale use. We report estimated marginal means per visualization from the mixed-effects model and their difference. The reductions in perceived temporal demand and effort persist under adjustment, confidence and success rate regret are unchanged.
  }
  \label{tab:study-adjusted}
  \small
  \begin{tabular}{@{}lrrl@{}}
  \toprule
  Outcome                      & Graph  & Videos & Difference [95\% CrI] \\
  \midrule
  Temporal demand $\downarrow$ & $38.7$ & $64.8$ & $-26.1$ $[-38.4, -14.0]$ \\
  Effort $\downarrow$          & $38.1$ & $61.0$ & $-22.9$ $[-37.7, -7.8]$ \\
  Confidence $\uparrow$        & $61.3$ & $58.8$ & $+2.5$ $[-9.8, +15.5]$ \\
  Success rate regret $\downarrow$ & $7.9$  & $7.0$  & $+0.9$ $[-6.1, +8.5]$ \\
  \bottomrule
  \end{tabular}
\end{table}

\subsection{Behavior Graphs for Progress Estimation (Q2)}
\label{sec:progress}

\subsubsection{Data, Baselines, and Metrics}

We evaluate behavior graphs as backbones for progress estimation via the methodology described in \Cref{subsec:use_cases}. We consider 43 rollout datasets, spanning tasks in simulation and the real world taken from LIBERO~\cite{liu_libero_2023}, Robomimic~\cite{mandlekar_robomimic_2021}, MimicGen~\cite{mandlekar_mimicgen_2023}, and tasks from the real-world DROID setup. LIBERO rollouts were obtained with \(\pi_{0.5}\)~\cite{pi05_2025} (multi-task datasets) and the rest of the rollouts were obtained with flow matching policies (single-task datasets), using observation encoder outputs as representation. We use 3-fold cross-validation over random train/val/test splits: graphs are fit on the train split, hyperparameters are optimized on the validation split, and metrics are computed on the unseen test split. Reported metrics are averaged over the three folds.

We compare our method against two state-of-the-art VLM-based progress estimators, Robometer~\cite{liang_robometer_2026} and TOPReward~\cite{chen_topreward_2026}, and against a baseline that fits a multilayer perceptron (MLP) on the same policy embeddings the graph is built on, which separates the contribution of the graph structure from that of the underlying representation. 

We evaluate the fidelity of the predicted progress curves and the accuracy of their final estimates. For fidelity, we compare predictions against a linear ground-truth progress ramp on successful rollouts: value-order correlation (VOC) and Kendall's \(\tau\) measure whether progress increases monotonically, while the mean absolute error (MAE) measures whether its values are correct. For final estimates, we report the area under the ROC curve (AUC) of classifying rollout outcomes from the final predicted progress.

\subsubsection{Results}

Averaged over all 43 datasets, behavior graphs achieve the best VOC, Kendall's $\tau$, and MAE (see \Cref{tab:progress-by-rubric}), outperforming in 3/4 domains on VOC, 4/4 on $\tau$, and 3/4 on MAE. The only domain where a baseline leads on these metrics is LIBERO, where the MLP fit on the same policy embeddings is slightly ahead on VOC and MAE. Baselines remain competitive on AUC, where Robometer is strongest on average. VLM-based models are capable of predicting outcomes, but predict progress with less fidelity. We demonstrate that given a trained policy and a dataset of successful trajectories, behavior graphs are a highly capable progress estimator without training any neural network and only minimal parameter fitting. Furthermore, outperforming the MLP baseline, which operates on the same policy embeddings that the graph is built on, demonstrates that the temporal and behavioral structure imposed by the behavior graph is a helpful prior that substantially improves over naive learning-based approaches, raising average VOC from 0.769 to 0.945.

\begin{table}
  \vspace{5pt}
  \centering
  \caption{\small Progress-estimation metrics over 43 datasets spanning 4 domains. AUC is undefined for the two Robomimic datasets in which every rollout succeeds (Can-PH, Lift-PH), so the Robomimic and Average AUC entries are over 8 and 41 datasets. On average, graph-based progress estimation outperforms all baselines in the fidelity of the prediction. In AUC, which measures success classification induced by the progress estimates, Robometer outperforms on average.}
  \label{tab:progress-by-rubric}
  \setlength{\tabcolsep}{3pt}
  \begin{tabular}{@{}llrrrr@{}}
  \toprule
  Domain (\#Datasets)                          & Method             & VOC $\uparrow$          & Kendall $\uparrow$         & MAE $\downarrow$           & AUC $\uparrow$\\
  \midrule
  \multirow{4}{*}{LIBERO (6)}        & Graph (ours) & 0.912          & \textbf{0.864} & 0.082          & 0.807 \\
                                     & MLP        & \textbf{0.940} & 0.861          & \textbf{0.068} & \textbf{0.864} \\
                                     & Robometer-4b       & 0.799          & 0.650          & 0.154          & 0.860 \\
                                     & TOPReward          & 0.433          & 0.329          & 0.259          & 0.664 \\
  \midrule
  \multirow{4}{*}{Real-world (19)} & Graph (ours) & \textbf{0.948} & \textbf{0.930} & \textbf{0.047} & \textbf{0.904} \\
                                     & MLP        & 0.731          & 0.564          & 0.173          & 0.782 \\
                                     & Robometer-4b       & 0.664          & 0.496          & 0.181          & 0.859 \\
                                     & TOPReward          & 0.736          & 0.572          & 0.192          & 0.674 \\
  \midrule
  \multirow{4}{*}{Robomimic (10)}    & Graph (ours) & \textbf{0.965} & \textbf{0.958} & \textbf{0.108} & 0.763 \\
                                     & MLP        & 0.682          & 0.565          & 0.198          & \textbf{0.871} \\
                                     & Robometer-4b       & 0.431          & 0.342          & 0.234          & 0.751 \\
                                     & TOPReward          & 0.478          & 0.359          & 0.308          & 0.767 \\
  \midrule
  \multirow{4}{*}{MimicGen (8)}      & Graph (ours) & \textbf{0.936} & \textbf{0.907} & \textbf{0.078} & 0.595 \\
                                     & MLP        & 0.841          & 0.688          & 0.125          & 0.742 \\
                                     & Robometer-4b       & 0.646          & 0.495          & 0.224          & \textbf{0.789} \\
                                     & TOPReward          & 0.603          & 0.464          & 0.235          & 0.724 \\
  \midrule
  \midrule
  \multirow{4}{*}{Average (43)}    & Graph (ours) & \textbf{0.945} & \textbf{0.923} & \textbf{0.072} & 0.802 \\
                                   & MLP        & 0.769          & 0.629          & 0.155          & 0.804 \\
                                   & Robometer-4b       & 0.625          & 0.482          & 0.197          & \textbf{0.824} \\
                                   & TOPReward          & 0.609          & 0.469          & 0.236          & 0.700 \\
  \bottomrule
  \end{tabular}
\end{table}


\subsection{Ablation Study (Q3)}
\label{sec:ablation}

We ablate the effect of several design decisions on graph construction quality across all three real-world tasks in our user study. We estimate ground truth subtask boundaries via gripper opening and closing as a heuristic. We split the datasets 60/20/20 for training, validation, and test splits and evaluate graph construction via the proxy task of subtask segmentation. Similarly to progress estimation, graphs are fit on train, hyperparameters (including the number of behaviors~\(K\)) are optimized on val, and metrics are reported on the unseen test split. We use two metrics: \(F_1\), computed at an intersection-over-union (IoU) threshold of 0.75, and mean IoU for predicted segments that have been matched to a ground truth counterpart.

\subsubsection{Results}

\Cref{tab:ablation} sweeps three embedding representations against the three dimensionality reduction techniques. Averaged over reductions, policy embeddings lead on both metrics, at $F_1$ of $0.226$ against $0.132$ for proprioception and $0.113$ for DINOv3, and the best single cell on each metric is a policy-embedding row. The reduction yields no clear quantitative result, with the best setting differing by representation. Qualitatively, we find that policy embeddings under UMAP produce the most legible graphs.

\subsubsection{Does graph creation work across policies and tasks?}

Graph creation produces qualitatively coherent graphs across six policy classes (diffusion~\cite{chi_diffusion_2023} and flow matching policies, $\pi_{0.5}$~\cite{pi05_2025}, OpenVLA~\cite{kim_openvla_2024}, $\pi_0$-FAST, Foundry-VLA) on both short- and long-horizon tasks in simulation and the real world. We provide the resulting graphs on the interactive website. As most datasets have no subtask annotations, we cannot quantitatively assess graph quality for all datasets.


\begin{table}[t]
\vspace{5pt}
\centering\small
\caption{\small Ablation study results for subtask segmentation. We report mean $\pm$ std over 3 datasets. Each \emph{mean} row averages every dataset's score over the three reductions, then reports the mean $\pm$ std of those averages across datasets. We report $F_1$ at 0.75 IoU threshold and the mean IoU over matched segments. On average, policy embeddings outperform proprioception and frozen DINOv3 features. While there is no clear quantitative conclusion for reductions, UMAP performs more consistently than the others and is what we use for our other experiments.}
\setlength{\tabcolsep}{4pt}
\begin{tabular}{@{}ll rr@{}}
\toprule
Representation & Reduction & $F_1(0.75)\uparrow$ & IoU$\uparrow$ \\
\midrule
\multirow{3}{*}{Proprioception}
  & none & 0.093{$\pm$0.075} & 0.523{$\pm$0.061} \\
  & PCA  & 0.120{$\pm$0.121} & 0.547{$\pm$0.098} \\
  & UMAP & 0.182{$\pm$0.022} & 0.598{$\pm$0.033} \\

  & \emph{mean} & 0.132 $\pm$ 0.072 & 0.556 $\pm$ 0.063 \\
\midrule
\multirow{3}{*}{Policy embeddings}
  & none & \textbf{0.264}{$\pm$0.108} & 0.633{$\pm$0.048} \\
  & PCA  & 0.254{$\pm$0.121} & \textbf{0.644}{$\pm$0.059} \\
  & UMAP & 0.160{$\pm$0.031} & 0.567{$\pm$0.022} \\
  & \emph{mean} & 0.226 $\pm$ 0.086 & 0.615 $\pm$ 0.032 \\
\midrule
\multirow{3}{*}{DINOv3}
  & none & 0.088{$\pm$0.030} & 0.462{$\pm$0.029} \\
  & PCA  & 0.082{$\pm$0.023} & 0.462{$\pm$0.022} \\
  & UMAP & 0.169{$\pm$0.086} & 0.566{$\pm$0.088} \\
  & \emph{mean} & 0.113 $\pm$ 0.019 & 0.497 $\pm$ 0.014 \\
\bottomrule
\end{tabular}
\label{tab:ablation}
\end{table}

\subsection{Discussion and Limitations}
 
On decision-making (Q1), behavior graphs compress deployment rollouts into a structure operators read faster, with less effort, and strongly prefer over rollout videos, while reaching comparable data collection decisions: neither confidence nor success rate regret differs reliably between the two. At the scale of our study, a well-informed operator reaches sound conclusions from videos alone, so the graph's observable value lies in how quickly they get there.
 
As a representation beyond interpretability (Q2), the same structure supports progress estimation: a piecewise-linear readout on the discovered behaviors outperforms both VLM-based and learned estimators on correlation and error, without requiring the subtask segmentation annotations that comparable stage-aware estimators depend on. Its main limitation is that the readout, like in other estimators, is non-causal and would need adapting for monitoring during inference.
 
On construction (Q3), the chosen representation for graph creation has a substantial impact on the results. Subtask segmentation is only a proxy, as interpretability is hard to evaluate, but the quantitative results favoring policy embeddings match our qualitative experience, typically leading to interpretable graphs. For the reduction, qualitative and quantitative findings are not always consistent: in practice, we use UMAP. The number of behaviors \(K\) also strongly shapes the graphs, though it is more of a user choice that sets their preferred level of detail. These limitations are largely related to subtask segmentation to construct graphs. We expect VLMs with better spatial understanding to annotate subtasks reliably and cheaply enough to replace clustering.

\section{Conclusion}

With behavior graphs, we aim to make large volumes of rollout data interpretable and actionable. Construction requires nothing beyond a trained policy and its rollouts, reusing representations the policy already provides. The resulting structures summarize policy behavior at a glance, even across hundreds of rollouts, and expose the failure modes that should drive the next round of data collection. In our study, operators reached the same conclusions from the graph as from rollout videos while spending markedly less perceived time and effort, and 16 of 18 chose it when given the option. Beyond serving as a visual summary, the structure imposed by behavior graphs benefits downstream tasks: a lightweight readout on the discovered behaviors estimates task progress more accurately than specialized VLM-based and learned estimators, without requiring annotations. We envision behavior graphs as a standard component of future robot data flywheels, giving operators a concrete, interpretable basis for acting on their deployment data.

\section*{Acknowledgments}
AI has been used to assist code implementation, figure creation, and text editing.
EB is supported by a Stanford Graduate Fellowship.
EB and CA are supported by Boeing under Grant No. 2026-UI-PA-172.
This research was supported in part by the Defense Advanced Research Projects Agency (DARPA) under Agreement No. HR00112490425. The views, opinions, and/or findings expressed are those of the author(s) and should not be interpreted as representing the official views or policies of the Defense Advanced Research Projects Agency (DARPA), or the U.S. Government.
Toyota Research Institute (TRI) provided funds to assist the authors with their research; this article solely reflects the opinions and conclusions of its authors and not TRI or any other Toyota entity.

\bibliographystyle{IEEEtran}
\bibliography{example}

\end{document}